\documentclass[10pt,twocolumn,letterpaper]{article}

\usepackage{iccv}
\usepackage{times}
\usepackage{epsfig}
\usepackage{graphicx}
\usepackage{amsmath}
\usepackage{amssymb}
\usepackage{wrapfig}
\usepackage{multirow}
\usepackage{makecell}
\usepackage{caption}
\usepackage{pifont}
\usepackage{xcolor}
\definecolor{darkgreen}{rgb}{0.0,0.6,0.0}
\usepackage[symbol]{footmisc}
\usepackage[linesnumbered,ruled]{algorithm2e}

\usepackage{bm}

\usepackage[pagebackref=true,breaklinks=true,letterpaper=true,colorlinks,bookmarks=false]{hyperref}

\newcommand{\cmark}{\textcolor{darkgreen}{\ding{51}}}
\newcommand{\xmark}{\textcolor{red}{\ding{55}}}

\usepackage{booktabs}
\usepackage{tabularx}
\usepackage{wrapfig}
\usepackage{enumitem}
\usepackage{xspace}

\usepackage[capitalize]{cleveref}
\crefname{section}{Sec.}{Secs.}
\Crefname{section}{Section}{Sections}
\Crefname{table}{Table}{Tables}
\crefname{table}{Tab.}{Tabs.}

\newcommand{\parahead}[1]{\noindent\textbf{#1}.\ }
\renewcommand{\paragraph}[1]{{\vspace{1mm}\noindent \bf #1}.}

\newcommand{\name}{COPILOT\xspace}

\usepackage[breaklinks=true,bookmarks=false]{hyperref}

\iccvfinalcopy %

\def\iccvPaperID{3436} %

\ificcvfinal\fi

\begin{document}

\title{COPILOT: Human-Environment Collision Prediction \\ and Localization from Egocentric Videos \vspace{-5mm}}
\author{Boxiao Pan$^{1}$\qquad Bokui Shen$^{1\footnote[1]{}}$\qquad Davis Rempe$^{1\footnote[1]{}}$ \qquad Despoina Paschalidou$^{1}$\\
\qquad Kaichun Mo$^{1,2}$ \qquad Yanchao Yang$^{1,3}$ \qquad Leonidas J. Guibas$^{1}$\\
\text{\normalsize $^1$Stanford University \qquad $^2$NVIDIA Research \qquad $^3$The University of Hong Kong}\\
}

\twocolumn[{
\renewcommand\twocolumn[1][]{#1}%
\maketitle
\vspace{-0.485in}
\begin{center}
    \centering
    \includegraphics[width=\textwidth]{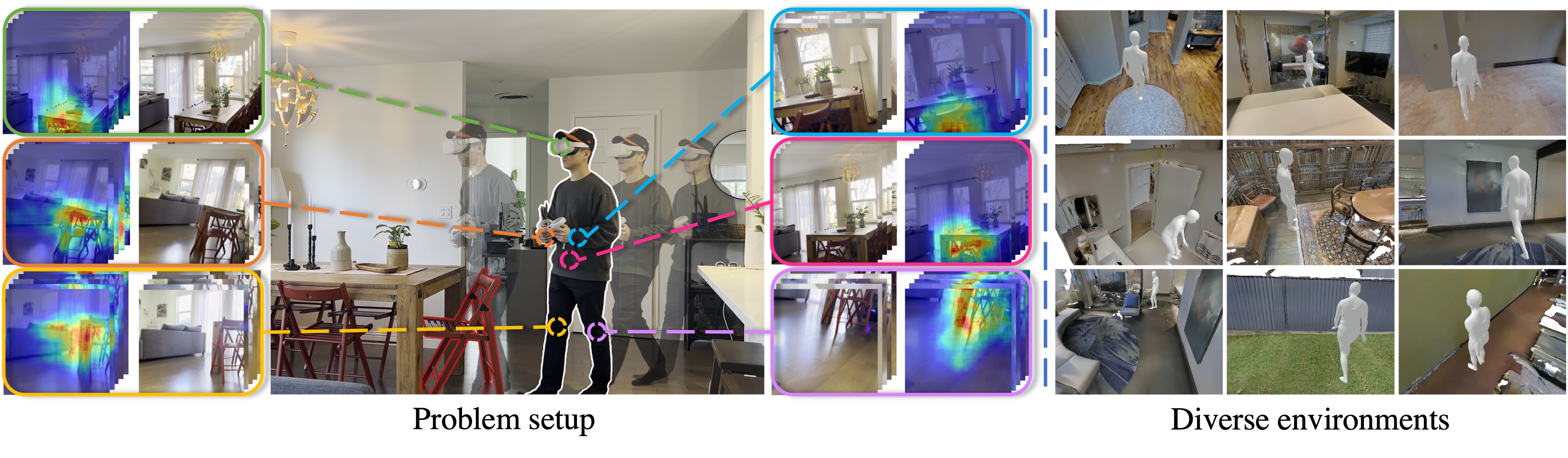}
    \captionof{figure}{\textbf{Problem formulation}. (Left) Egocentric videos from cameras mounted on the body are used to perform three tasks: binary collision prediction, classifying the colliding joints, and localizing collision regions in the environment. In this example, our model predicts and localizes a collision between the left knee and the chair. (Right) Our model is trained on a newly proposed large-scale synthetic dataset and generalizes to diverse synthetic and real-world environments.}
    \label{fig:problem_setup}
\end{center}%
}]
\maketitle
\ificcvfinal\thispagestyle{empty}\fi

\footnotetext[0]{* Equal contribution}

\begin{abstract}
    \vspace{-4mm}
    The ability to forecast human-environment collisions from egocentric observations is vital to enable collision avoidance in applications such as VR, AR, and wearable assistive robotics. 
    In this work, we introduce the challenging problem of predicting collisions in diverse environments from multi-view egocentric videos captured from body-mounted cameras.
    Solving this problem requires a generalizable perception system that can classify which human body joints will collide and estimate a collision region heatmap to localize collisions in the environment.
    To achieve this, we propose a transformer-based model called COPILOT to perform collision prediction and localization simultaneously, which accumulates information across multi-view inputs through a novel 4D space-time-viewpoint attention mechanism.
    To train our model and enable future research on this task, we develop a synthetic data generation framework that produces egocentric videos of virtual humans moving and colliding within diverse 3D environments. This framework is then used to establish a large-scale dataset consisting of 8.6M egocentric RGBD frames.
    Extensive experiments show that COPILOT generalizes to unseen synthetic as well as real-world scenes. We further demonstrate COPILOT outputs are useful for downstream collision avoidance through simple closed-loop control. 
    Please visit our project webpage at \url{https://sites.google.com/stanford.edu/copilot}.
\end{abstract}

\vspace{-6mm}
\section{Introduction}
Forecasting potential collisions between a human and their environment from \textit{egocentric} observations is crucial to applications such as virtual reality (VR), augmented reality (AR), and wearable assistive robotics.
As users are immersed in VR / AR, the effective and non-intrusive prevention of collisions with real-world surroundings is critical.
On the other hand, wearable exoskeletons~\cite{ExoClassical2007,chen2017wearable} provide physical assistance to users with mobility impairments and therefore must control user movements to avoid dangerous falls by estimating \textit{when} and \textit{where} a collision may occur.

To this end, a multitude of works have explored ways to predict and localize future collisions. For example, existing approaches for VR leverage redirected walking techniques \cite{RDW2005} to guide users in the virtual world such that they avoid collisions with real-world objects~\cite{CollisionVR2012, SLAMVR2018, APFRDW2019, DynamAPF2020, VisPolyVR2021}.
However, these approaches rely on pre-scanned scenes and external tracking systems \cite{CollisionVR2012, APFRDW2019, DynamAPF2020, VisPolyVR2021} or resort to online SLAM \cite{SLAMVR2018, VRoamer2019}, making them cumbersome and tedious to use.
For assistive exoskeletons, several methods have been proposed to classify environment types \cite{ExoCNNEnvCls2022, ExoCNNStairCls2022}, recognize obstacle parameters \cite{ExoDepthObjRec2019, ExoDepthObjRec2020}, predict collision risks \cite{ExoMotionCls22020, ExoMotionCls2020}, and predict future trajectories \cite{ExoGaze2020, ExoFlowJointPred2022}. 
While these works are useful in constrained settings, they impose several constraints to simplify the problem, such as assuming pre-defined environments \cite{ExoMotionCls2020, ExoDepthObjRec2020, ExoCNNStairCls2022} or performing only environment classification rather than explicit collision forecasting \cite{ExoDepthTerrainCls2021, ExoCNNEnvCls2022, ExoRGBDEnvCls2022, ExoEnvClsSim2Real2022}.

In this work, we address the shortcomings of prior approaches to collision forecasting by introducing a generalizable model that provides detailed perception outputs useful for downstream collision avoidance. %
We first formalize the problem of egocentric collision prediction and localization from single or multi-view videos captured by cameras mounted on a person. 
To tackle this problem, we design a multi-task architecture that scales to diverse and complex environments and provides rich collision information. %
We train our model using high-quality simulation and curate data from a large variety of synthetic environments.

We pose collision forecasting as a problem of \textit{classifying} which body joints will collide in the near future and \textit{localizing} regions in the input videos where these collisions are likely to occur. 
The input is a set of multi-view RGB egocentric videos (see \cref{fig:problem_setup}) \emph{without} camera poses, which makes our problem setup general but also more challenging.
To successfully predict when and what part of the human body will collide with the environment, we must accumulate information about the human's movement, intent, and their surroundings across the multi-view videos.

To address these challenges, we propose \name, a \textbf{CO}llision \textbf{P}red\textbf{I}ction and \textbf{LO}calization \textbf{T}ransformer that uses a novel 4D attention scheme across space, time, and viewpoint to leverage information from multi-view video inputs. 
By alternating between cross-viewpoint and cross-time attention, the proposed 4D attention fuses information on both scene geometry and human motion.
Notably, \name classifies collisions and estimates collision regions simultaneously in a multi-task fashion, thereby improving performance on collision prediction and providing actionable information for collision avoidance as well.

To train and evaluate our model, we develop a data framework and curate a high-quality synthetic dataset containing $\sim$8.6M egocentric RGBD frames with automatically-annotated collision labels and heatmaps. 
Data is generated in simulation~\cite{Habitat2019, Habitat2021} using real-world scene scans~\cite{Matterport3d2017,Gibson2018} and a human motion generation model pre-trained with motion capture data \cite{Humor2021}. Our data framework and dataset feature large scene diversity, realistic human motion, and accurate collision checking. 
Experiments show that \name achieves over 84\% collision prediction accuracy on unseen synthetic scenes after training on our large-scale dataset, and qualitatively generalizes to diverse real-world environments.
To mimic the exoskeleton application, we combine our model with a closed-loop controller in simulation to demonstrate that we can adjust human motion for improved collision avoidance; this avoids 35\% and 29\% of collision cases on training and unseen synthetic scenes, respectively.

In summary, we make the following \textbf{contributions}: (1) we introduce the challenging task of collision prediction and localization from unposed egocentric videos. (2) We propose a multi-view video transformer-based model that employs a novel 4D space-time-viewpoint attention mechanism to solve this task effectively. (3) We develop a synthetic data framework to generate a large-scale dataset of egocentric RGBD frames with automatically-annotated collision labels and heatmaps. Our model trained on this dataset is shown to generalize to diverse synthetic and real-world environments through extensive evaluations.
The dataset and code will be made available upon acceptance.

\section{Related Work}

\parahead{Collision perception for VR}
To avoid potential user collisions with the real-world environment, existing works combine Redirected Walking (RDW) \cite{RDW2005} with techniques such as Artificial Potential Fields \cite{APFRDW2019, DynamAPF2020} and Reinforcement Learning \cite{QRDW2019, RLRDW2021}. On a high level, RDW maps the obstacles in the physical world to some form in the virtual world to imperceptibly steer the user along a collision-free path. These techniques require known scene geometries and user locations in the physical world, which normally come from scene scanning and external tracking systems. Some works try to relax these assumptions by fusing depth observations online \cite{SLAMVR2018, VRoamer2019}, but they require additional hardware and tedious post-processing. %
As our approach directly predicts collisions from egocentric video streams, it does not require cumbersome external hardware setups.

\parahead{Collision perception for exoskeletons}
Assistive wearable exoskeletons provide physical assistance to modify a user's actions and avoid collisions with the environment.
To achieve this, environment perception is necessary. Prior works classify environments into one of several pre-defined categories \cite{ExoCNNEnvCls2022, ExoRGBDEnvCls2022, ExoCNNStairCls2022}, recognize obstacle parameters \cite{ExoDepthObjRec2019, ExoDepthObjRec2020}, or predict short-term future user motions \cite{ExoMotionCls2020, ExoGaze2020, ExoFlowJointPred2022}. 
These estimations are fed to hand-designed algorithms to guide the behavior of the exoskeleton controller. In contrast, our approach explicitly forecasts collisions, without relying on any post-processing.
Moreover, these works operate in simple environments with few obstacles and only consider the lower body limbs, unlike \name that generalizes to diverse environments and predicts full-body collisions.

\parahead{Perception from egocentric videos}
While egocentric vision spans a diverse range of topics, the most relevant to our work are body pose estimation, human-object interaction prediction, and collision prediction. 
Estimating and forecasting full-body poses from egocentric videos is an under-constrained problem since multiple human poses can plausibly explain the same egocentric observations. Prior works use a downward-facing fish-eye camera to better capture body parts \cite{XREgoPose2019, GlobalEgoPose2021}, regularize estimations with dynamics and kinematics models \cite{RLEgoPose2019, DynamicsEgoPose2021}, or leverage multi-person interaction cues in the egocentric view \cite{You2Me2020}.
In this paper, we do not focus on this task but instead seek to exploit cues from the input egocentric videos to reason about potential collisions with the environment in the near future.
Other works infer human-object interactions from egocentric videos, such as future 2D hand trajectories and interaction hotspots \cite{FutureHoI2020, InteractHotspot2022}. 
The closest to ours is \cite{KitaniCollision2019}, which predicts the time-to-environment collision for a suitcase-shaped robot, given only a single-view RGB video. %
In contrast, we predict fine-grained per-joint collisions and collision heatmaps for the more complex human morphology.

\parahead{General video understanding}
Spatio-temporal reasoning is one of the main topics for video understanding. 
Earlier approaches perform 3D convolutions to allow hierarchical spatio-temporal understanding \cite{C3D2015,I3D2017,TwoStream2016,NonLocal2018,SlowFast2019,FeatureBank2019}. Recently, several works \cite{ViViT2021, MViT2021, VideoSwin2021, Timesformer2021, RevT2022} have adapted powerful transformer architectures \cite{Transformer2017} to the video domain. Similar to ViT \cite{ViT2020}, TimeSformer \cite{Timesformer2021} divides video frames into patches and proposes several strategies to perform spatio-temporal self-attention. 
This architecture achieves strong performance on capturing spatio-temporal patterns in video data, but only handles a single video stream. 
Our backbone extends TimeSformer to use a novel 4D attention operating over the space-time-viewpoint domain.

\begin{figure*}[t]
    \centering
    \includegraphics[width=\linewidth]{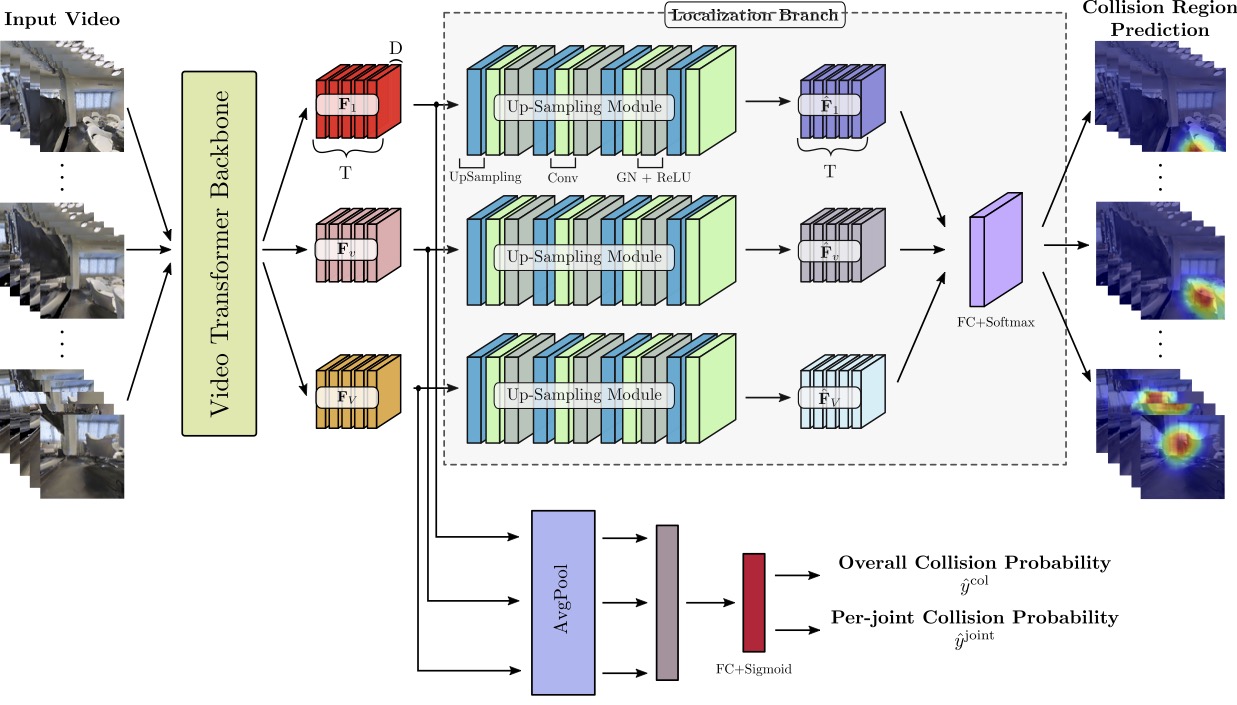}
    \caption{
    \textbf{\name architecture}. Videos taken from different viewpoints are fed to a backbone transformer model that uses 4D space-time-viewpoint attention to output a set of feature maps. The localization branch gradually up-samples the features to the original image resolution via the up-sampling module, which are then used to estimate the collision region heatmaps. The prediction branch pools the features globally to predict the overall and joint-level collisions.
    }
    \vspace{-3mm}
    \label{fig:framework}
\end{figure*}
\section{Problem Formulation and Data Generation}
\subsection{Collision Prediction and Localization}
\label{sec:formulation}
To provide useful information for downstream collision avoidance, we need to know \textit{if} there will be a collision between the person and the environment as well as \textit{where} this collision will occur, both on the human body and in the scene. %
Given multi-view egocentric videos, we formulate this problem as (1) classifying whether a collision will happen in the near future, (2) identifying the body joints involved in a potential collision, and (3) localizing the collision region in the scene.
Unlike prior works that investigate collisions only for lower limbs~\cite{ExoCNNEnvCls2022,ExoDepthObjRec2020,ExoFlowJointPred2022,ExoClassical2007,ExoCNNStairCls2022,ExoMotionCls2020,ExoGaze2020}, we expand our scope to full-body collision prediction
that utilizes multi-view inputs from cameras mounted on several body joints.
Moreover, as camera extrinsics and intrinsics can be hard to acquire for in-the-wild scenes, we rely solely on visual inputs. 
These unposed multi-view input videos have the potential to capture the full-body motion and environment context better than a single view.

Given a set of multi-view, time-synchronized egocentric RGB videos $X = \{X_1, X_2, \ldots, X_V\}$ captured from $V$ viewpoints where $X_v = \{x^v_1, x^v_2, \ldots, x^v_T\}$ consists of $T$ frames captured from the $v$-th camera, our goal is to predict:
\begin{enumerate}[leftmargin=*,itemsep=0pt]
    \item An \textit{overall} binary prediction $y^{\text{col}}$ indicating whether any collision will happen in the next $H$ time steps. This determines \textit{whether} to intervene to avoid a collision;
    \item A \textit{per-joint} binary prediction $y^{\text{joint}}_j$ indicating whether the $j$-th joint will be involved in a collision. Note that typically,
    the number of joints $J$ is different from the number of views $V$, because attaching cameras to all body joints can be impractical. This indicates \textit{where} to intervene;
    \item A set of per-frame collision region heatmaps $y^{\text{map}} = \{P_1, P_2, \ldots, P_V\}$, where $P_v = \{p^v_1, p^v_2, \ldots, p^v_T\}$ is the heatmap sequence corresponding to $X_v$. Note that the sum of all values in one heatmap is one, namely $\sum_{ij} p^v_{tij} = 1$, so that $p^v_t$ is a probability distribution over all spatial locations in that frame. This represents the likelihood that a collision will occur at each location and indicates \textit{how} to intervene. 
\end{enumerate}

\noindent %
Although in the above formulation we only consider RGB inputs, leveraging depth data is also possible.
However, we choose RGB images as our primary setting because they are easier to acquire than depth images, which rely on cumbersome depth cameras that are not always available. 

\subsection{Large-Scale Data Generation}
\label{sec:datagen}
Due to the anomalous nature of collision events, it is unclear how to collect real-world data for our task, let alone fine-grained annotations for collision regions and colliding joints. Hence, we build a framework that leverages a simulated environment to generate a large-scale high-quality synthetic dataset containing $\sim$8.6M egocentric RGBD frames with automatically-annotated collision labels and heatmaps. We describe the key steps of this framework next and leave the details to the supplementary material. 

Our data framework places virtual humans in synthetic scenes and renders egocentric videos as they perform realistic motions.
To get a diverse set of environments, we pick 100 scenes from the Matterport3D \cite{Matterport3d2017} and Gibson \cite{Gibson2018} datasets, where each scene contains several rooms.
We leverage HuMoR \cite{Humor2021} to randomly generate sequences of realistic walking motions in each scene using the SMPL human body model~\cite{SMPL2015}.
During sequence rollout, we perform collision checking between the human and scene meshes to get the \textit{overall} collision label $y^{\text{col}}$. We then assign each vertex where a collision occurred to the closest one of 10 human body joints (head, torso, elbows, hands, legs, and feet), which results in a set of \textit{per-joint} collision labels $y^{\text{joint}}$. 
Collision checking is efficiently implemented using PyBullet \cite{PyBullet2021} by segmenting the body into convex parts; when a collision is detected, the sequence is terminated.

For each generated motion sequence, observations are rendered from egocentric viewpoints with the AI-habitat simulator \cite{Habitat2021,Habitat2019} by placing cameras at 6 body joints (head, pelvis, wrists, and knees). %
Each sequence is split into subsequences of one second length that are used as input to our model; each subsequence is labeled as ``colliding'' if a collision is detected within one second in the future (\ie $H = 30$ at 30 fps).
We empirically found (\cref{subsec:control_exp}) that videos of a one-second duration typically have enough evidence to decide whether to act to avoid a potential collision. %
Finally, to generate the collision region heatmaps $y^{\text{map}}$, we project the vertices in the scene that the person collided with to each egocentric viewpoint, and set these pixels to 1 and the rest to 0. We apply a Gaussian kernel to smooth the heatmap before normalizing it to a proper distribution.
\section{Method}

Our goal is to develop a perception module that can jointly solve the collision prediction and collision region localization tasks.
As these tasks are deeply connected and both require learning features of motion and environment geometry, we introduce \name to tackle them jointly in a multi-task fashion (supported experimentally in \cref{para:multi-task}).
As shown in \cref{fig:framework}, the input egocentric videos are first fed into a transformer encoder, which performs attention over space, time, and viewpoint across all video streams using our novel 4D attention mechanism.
Next, the \textit{localization branch} takes the output features and estimates collision region heatmaps for each viewpoint, while the \textit{prediction branch} performs global pooling
on the features and predicts the overall and per-joint collision probabilities. %
\begin{figure}[t]
    \centering
    \vspace{-4mm}
    \includegraphics[width=0.95\linewidth]{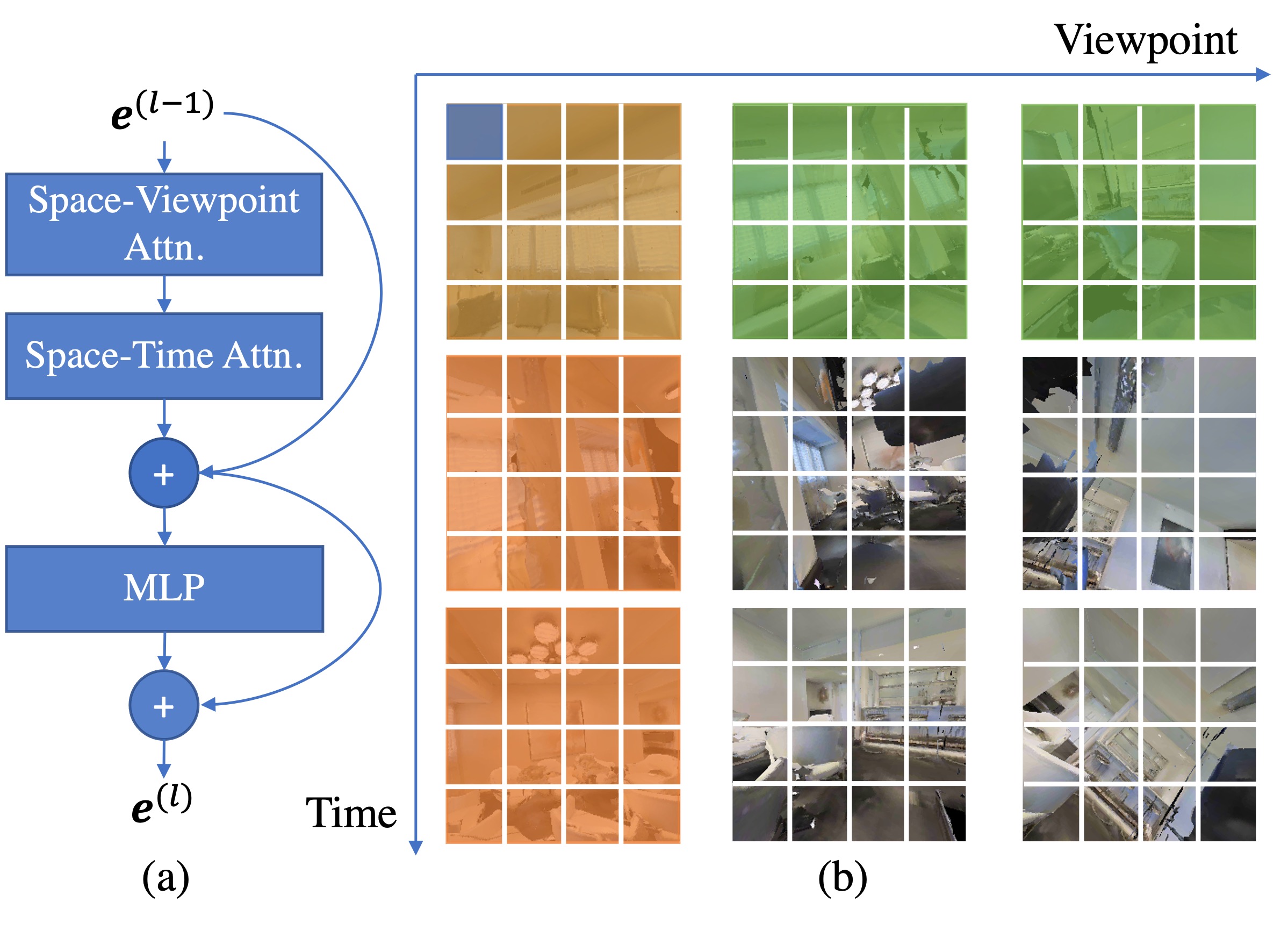}
    \vspace{-3mm}
    \caption{
    \textbf{4D STV attention}. (a) Attention block overview. (b) Illustration of the attention operation. Patches from three time steps and viewpoints are visualized. Green and orange denote patches that the blue patch attends to along the space-viewpoint and space-time dimensions, respectively. The current frame (upper left) has both colors.
    }
    \label{fig:attention}
    \vspace{-0.8em}
\end{figure}

\subsection{Transformer Backbone with 4D Attention}
Our backbone is based on TimeSformer~\cite{Timesformer2021}, which we extend to process videos from several views jointly using cross-viewpoint attention. Following the original architecture~\cite{Timesformer2021}, the model first splits each $H{\times}W$ RGB frame into $N$ non-overlapping spatial patches, each of size $P{\times}P$. %
Each patch is flattened into a vector $\displaystyle \mathbf{z}_{(p, t, v)} \in \mathbb{R}^{F}$, where $p$, $t$, and $v$ index space, time, and viewpoint, respectively, and $F=3 \cdot P \cdot P$.
Each vector is then mapped to a \textit{patch embedding} $\mathbf{e}_{(p, t, v)}^{(0)} \in \mathbb{R}^{D}$ as

\begin{equation}
    \mathbf{e}_{(p, t, v)}^{(0)} = E \mathbf{z}_{(p, t, v)} + \text{emb}_{(p, t, v)}^{\text{pos}},
\end{equation}
where $E \in \mathbb{R}^{D \times F}$ is a learnable matrix and $\text{emb}_{(p, t, v)}^{\text{pos}} \in \mathbb{R}^D$ is a learnable positional embedding. These patch embeddings are the input to the transformer encoding blocks.

In particular, for a single encoding block $l \in [1, L]$ (see Fig.~\ref{fig:attention}), the query vectors
are computed as %
\begin{equation}
\mathbf{q}_{(p, t, v)}^{(l, a)} = W_{Q}^{(l, a)} \texttt{LN} \left(\mathbf{e}_{(p, t, v)}^{(l - 1)} \right),
\end{equation}
where \texttt{LN()} denotes LayerNorm \cite{LayerNorm2016} and $a \in [1, \mathcal{A}]$ indexes the attention heads. %
Similarly, we compute the keys $\mathbf{k}_{(p, t, v)}^{(l, a)}$ and values $\mathbf{v}_{(p, t, v)}^{(l, a)}$.

TimeSformer \cite{Timesformer2021} proposes several attention mechanisms to perform space-time self-attention on a single video stream. One way to adapt their model to our multi-view setting is to perform space-time attention for each stream separately and then fuse the resulting features with an MLP (\ie late feature fusion).
However, this strategy does not fully exploit the overlapping multi-view information relevant to scene geometry and human motion. We thus propose a 4D space-time-viewpoint (STV) attention operation, which is illustrated in \cref{fig:attention}. For each patch, the model first attends to all patches across different viewpoints at the same time step (\ie space-viewpoint attention), then attends to all patches within the same viewpoint (\ie space-time attention). Intuitively, space-viewpoint attention gathers environment information from all viewpoints, while space-time attention captures dynamics from a single viewpoint. 

Formally, we first compute self-attention weights $\boldsymbol{\alpha}_{\text{SV}} \in \mathbb{R}^{N\cdot V}$ and $\boldsymbol{\alpha}_{\text{ST}} \in \mathbb{R}^{N\cdot T}$ for each attention type:

\begin{align}
    \boldsymbol{\alpha}_{\text{SV} (p, t, v)}^{(l, a)} &= \texttt{SM} \left(\frac{{\mathbf{q}_{(p, t, v)}^{(l, a)}}^\intercal}{\sqrt{D_h}} \cdot \left\{ \mathbf{k}_{(p', t, v')}^{(l, a)} \right\}_{\substack{p' = 1, \ldots, N \\ v' = 1, \ldots, V}} \right) \\
    \boldsymbol{\alpha}_{\text{ST} (p, t, v)}^{(l, a)} &= \texttt{SM} \left(\frac{{\mathbf{q}_{(p, t, v)}^{(l, a)}}^\intercal}{\sqrt{D_h}} \cdot \left\{ \mathbf{k}_{(p', t', v)}^{(l, a)} \right\}_{\substack{p' = 1, \ldots, N \\ t' = 1, \ldots, T}} \right)
\end{align}
where \texttt{SM()} denotes the softmax function. The output feature encoding $\mathbf{e}_{(p, t, v)}^{(l)}$ is computed by first summing the value vectors with the attention weights:
\begin{align}
    \mathbf{s}_{(p, t, v)}^{(l, a)} = 
    &\sum_{p'=1}^{N} \sum_{v'=1}^{V} \alpha_{\text{SV} (p, t, v),(p', v')}^{(l,a)} \mathbf{v}_{(p', v', t)}^{(l, a)} \\
    + &\sum_{p'=1}^{N} \sum_{t'=1}^{T} \alpha_{\text{ST} (p, t, v),(p', t')}^{(l,a)} \mathbf{v}_{(p', v, t')}^{(l, a)}.
\end{align}
Then, the encodings from all heads are concatenated and mapped through an MLP with residual connections to obtain the final feature output of the block:
\begin{align}
    \mathbf{e'}_{(p, t, v)}^{(l)} &= W_O \left[ \mathbf{s}_{(p, t, v)}^{(l, 1)} \ldots \mathbf{s}_{(p, t, v)}^{(l, \mathcal{A})} \right] + \mathbf{e}_{(p, t, v)}^{(l - 1)} \\
    \mathbf{e}_{(p, t, v)}^{(l)} &= \texttt{MLP} \left( \texttt{LN} \left( \mathbf{e'}_{(p, t, v)}^{(l)} \right )\right) + \mathbf{e'}_{(p, t, v)}^{(l)}.
\end{align}

\subsection{Collision Region Localization}
\label{subsec:col_region_method}
The output feature grid is then passed to the \textit{localization branch}, which predicts the per-frame spatial heatmaps corresponding to the input videos.
High values in the heatmap indicate areas in the scene where there is a high probability for a collision to occur in the near future.
Inferring spatial collision heatmaps requires per-pixel predictions, so this branch of the model must up-sample the discretized feature maps back to the original image resolution. Inspired by the scheme introduced in FPN \cite{FPN2017}, we use a sequence of up-sampling modules, each doing $2\times$ nearest-neighbor up-sampling followed by a $3{\times}3$ convolution, GroupNorm layer~\cite{wu2018group}, and ReLU activation. Different from  \cite{FPN2017}, the feature dimension is reduced by half in each spatial up-sampling to reduce the memory allocation. 
Finally, we use a one-layer MLP followed by a \textit{softmax} to get the collision region heatmaps $\hat{y}^{\text{map}}_v \in [0, 1]^{H\times W}$, such that each heatmap is a distribution of collision probability:
\begin{equation}
    \hat{y}^{\text{map}}_v = \texttt{SM}\left(\texttt{MLP}\left(\texttt{up}\left(\mathbf{e}_{(t, v)}^{(L)}\right)\right)\right),
\end{equation}
where $\mathbf{e}_{(t,v)}^{(L)}$ is the $(t, v)$-th slice of $\mathbf{e}^{(L)}$.

\subsection{Overall and Per-Joint Collision Prediction}
\label{subsec:col_prediction_method}
Likewise, we pass the output features from our transformer backbone to the \textit{prediction branch} that predicts: (1) the \textit{overall} binary collision probability,
namely whether the person will collide within the next $H$ timesteps, and (2) the \textit{per-joint} binary collision probability indicating whether each joint will collide.
Features are first pooled globally using average-pooling across all space-time-viewpoint.
Then, this feature vector is fed to an MLP followed by a sigmoid activation to estimate the overall and per-joint collision predictions as follows:
\begin{equation}
    \hat{y}^{\text{joint}}, \hat{y}^{\text{col}} = \texttt{Sig}\left(\texttt{MLP}\left(\texttt{pool}\left(\mathbf{e}^{(L)}\right)\right)\right) \in [0, 1]^{J + 1},
\end{equation}
where the first $J$ outputs from the MLP are used for joint collision prediction $\hat{y}^{\text{joint}} \in [0, 1]^{J}$ while the last value is used for overall collision prediction $\hat{y}^{\text{col}} \in [0, 1]$.

\subsection{Training}
\name is trained in a fully-supervised manner using the synthetic dataset described in Sec.~\ref{sec:datagen}. The overall loss is the sum of three terms
\begin{equation}
    L = L_{\text{map}} + L_{\text{col}} + L_{\text{joint}},
\end{equation}
where $L_{\text{map}} = D_{\text{KL}} (\hat{y}^{\text{map}} || y^{\text{map}})$ is the KL-divergence between the predicted and the ground-truth collision heatmaps,
$L_{\text{col}} = \text{BCE} (\hat{y}^{\text{col}}, y^{\text{col}})$ is the binary cross-entropy loss (BCE) between the predicted and the target overall collision label for the motion sequence, and $L_{\text{joint}} = \sum_{j=1}^J \text{BCE} (\hat{y}^{\text{joint}}_j, y^{\text{joint}}_j)$ is the sum of the binary cross-entropy losses between the target and the
predicted per-joint collision probabilities.

\paragraph{Implementation Details}
We use PyTorch Lightning \cite{PL2019} and train using SGD~\cite{SGD2016} with an initial learning rate of 1e-4, which is decayed by $0.1$ at the $25$-th, $35$-th, and $45$-th epochs. All training is done on a single Nvidia RTX A5000 GPU, and the model requires training for approximately two days to converge.
For all experiments, the model is given 1 sec of past observations (sampled at 5 Hz) and predicts collisions up to 1 sec into the future.

\section{Experimental Results}
\label{sec:experiments}
We compare \name to baselines in both single-view and multi-view settings in Sec.~\ref{subsubsec:attn}. In Sec.~\ref{subsubsec:ablate}, we investigate the impact of various components of \name on its performance. We then present qualitative evaluation results of collision region localization, both in simulation (Sec.~\ref{subsec:qualitative}) and the real world (Sec.~\ref{subsec:real_world_exp}). Additionally, we show that outputs from \name are useful by applying \name to a collision avoidance task. The supplementary material provides additional details of experiments and metrics, along with additional visualizations and analyses.

\begin{table*}[t!]
    \centering
    {%
    \setlength{\tabcolsep}{4pt}
    \begin{tabular}{lcl|ccccc|ccccc}
        \toprule
        & \multicolumn{1}{c}{} & \multicolumn{1}{c|}{} & \multicolumn{5}{c|}{\textbf{Unseen Motions}} & \multicolumn{5}{c}{\textbf{Unseen Motions \& Scenes}} \\
        \textbf{Method} & \textbf{Views} & \multicolumn{1}{c|}{\textbf{Attention}} & Col & F1@4 & F1@6 & F1@10 & KL & Col & F1@4 & F1@6 & F1@10 & KL \\
        \midrule
        Time2Col~\cite{KitaniCollision2019} & Root & \multicolumn{1}{c|}{-} & 61.9 & 56.3 & 43.9 & \textbf{38.0} & 0.86 & 64.3 & 47.3 & 33.9 & 29.4 & \textbf{0.94} \\
        \name & Root & 3D ST \cite{Timesformer2021} & \textbf{74.5} & \textbf{57.1} & \textbf{44.7} & 37.9 & \textbf{0.79} & \textbf{77.1} & \textbf{56.0} & \textbf{40.7} & \textbf{33.2} & 0.99 \\
        \midrule
         &  & 3D ST \cite{Timesformer2021} & 80.9 & 69.1 & 55.7 & 49.6 & 0.95 & \textbf{84.8} & 57.4 & 42.5 & 37.0 & 1.41 \\
        \name & All  & 4D STV Div & 80.0 & 66.8 & 57.7 & 52.3 & 0.82 & 81.0 & 53.7 & 43.7 & 37.1 & 1.25 \\
        &  & 4D STV Full (ours) & \textbf{86.0} & \textbf{71.2} & \textbf{62.2} & \textbf{55.6} & \textbf{0.64} & \textbf{84.8} & \textbf{63.7} & \textbf{52.9} & \textbf{45.6} & \textbf{1.09} \\
        \bottomrule
    \end{tabular}
    }
    \vspace{-2mm}
    \caption{\textbf{Comparison with baselines.} (Top) Our transformer-based backbone outperforms a convolutional baseline~\cite{KitaniCollision2019}. (Bottom) Our proposed 4D attention scheme outperforms alternative approaches to multi-view fusion. All variants perform both collision prediction and localization in a multi-task manner. Classification accuracies and F1 scores are percentages (\%).
    }
    \label{table:quant_eval}
    \vspace{-3mm}
\end{table*}

\begin{table}[t]
    \centering
    {%
    \setlength{\tabcolsep}{2pt}
    \begin{tabular}{ccc|cc|cc}
        \toprule
        \multicolumn{2}{c}{\textbf{\small Input}} & \textbf{\small Output} & \multicolumn{2}{c|}{\textbf{\small Unseen M.}} & \multicolumn{2}{c}{\textbf{\small Unseen M. \& S.}} \\
        Views & Modality & Map & Col & F1@10 & Col & F1@10\\
        \midrule
        Root & RGB & \xmark & 69.2 & \textbf{38.7} & 69.9 & 32.4 \\
        Root & RGB & \cmark & \textbf{74.5} &  37.9 & \textbf{77.1} & \textbf{33.2} \\
        \midrule
       All & RGB & \xmark & 70.7 & 43.0 & 74.7 & 33.0 \\
        All & RGB & \cmark & \textbf{86.0} & \textbf{55.6} & \textbf{84.8} & \textbf{45.6} \\
        \midrule
       All & Depth & \cmark & 85.0 & 64.7 & 87.9 & 60.1 \\
        \bottomrule
    \end{tabular}
    }
    \vspace{-1mm}
    \caption{\textbf{Ablation study.} Our \name model achieves improved performance when predicting collision heatmaps in addition to classification (multi-task), using multi-view inputs, and using alternative input modalities such as depth.
    }
    \vspace{-4mm}
    \label{table:ablation}
\end{table}

\subsection{Quantitative Evaluation}
\label{subsec:quantitative}
We evaluate \name and baselines on unseen motion sequences from the synthetic dataset described in Sec.~\ref{sec:datagen}. 
We consider two evaluation setups. In the first, we evaluate on training scenes but novel motion sequences, whereas in the second, both scenes and motion sequences are not seen during training. We refer to the former as \textit{Unseen Motions} and to the latter as \textit{Unseen Motions \& Scenes}.
To measure the quality of the overall collision prediction $\hat{y}^{\text{col}}$, we report the binary classification accuracy (\textit{Col}). 
For the per-joint collision predictions $\hat{y}^{\text{joint}}$, we report F1-score (\textit{F1@N, N=4,6,10}) at different joint-level ``resolutions''. Specifically, we group the 10 individual joints predicted by \name into 4 and 6 groups based on semantics (\eg elbow and hand joints grouped together), and report the F1-score for each group. This provides multiple evaluation granularities since requirements may differ across various applications.
Finally, collision region localization $\hat{y}^{\text{map}}$ is evaluated in terms of the KL-divergence (\textit{KL}) between predicted and ground truth heatmaps. 

\vspace{-2mm}
\subsubsection{Comparison with Baselines}
\vspace{-2mm}
\label{subsubsec:attn}
We evaluate our model against existing backbone models and attention schemes for both the collision prediction and localization tasks and summarize the results in \cref{table:quant_eval}.
Since we study a novel problem, there is no directly comparable prior work.  Hence, we adapt \textit{Time2Col}~\cite{KitaniCollision2019}, which performs 2D convolutions on each frame and concatenates features for all frames to perform both collision prediction and localization. Our transformer-based backbone outperforms the convolutional approach of \textit{Time2Col} given a single viewpoint as input (from the root joint, \ie pelvis). 

Under the multi-view setting, we compare two ways of fusing the multi-view information: (1) 3D spatio-temporal attention (\textit{3D ST}) \cite{Timesformer2021}, which performs self-attention within each viewpoint, and concatenates features from all viewpoints for prediction, and (2) 4D divided attention over space-time-viewpoint (\textit{4D STV Div}), which extends divided space-time attention \cite{Timesformer2021} by attending only to patches at the same spatial location across viewpoint and time, followed by a spatial self-attention over its own frame. Our proposed space-time-viewpoint attention scheme (\textit{4D STV Full}) outperforms these baselines, due to its larger receptive field and fusing of cross-view information at an earlier stage.

\vspace{-2mm}
\subsubsection{Ablation Study}
\label{subsubsec:ablate}
We ablate important design decisions of \name and summarize the results in Tab.~\ref{table:ablation}. 

\paragraph{Multi-task learning}
\label{para:multi-task}
First, we evaluate whether performing both collision prediction and localization in a multi-task fashion improves performance. Models supervised with heatmap prediction are indicated in the \textit{Map} column. As shown, training in this multi-task setting that includes both collision prediction and collision region localization improves prediction performance for both the overall and the per-joint collisions. This trend holds for both single-view and multi-view settings, and the effect is most significant in unseen scenes (right section of \cref{table:ablation}).
We hypothesize that collision region localization encourages better scene understanding, which helps with cross-scene generalization. 

\paragraph{Multi-view inputs} 
We next investigate if using multi-view inputs indeed benefits the prediction task. In our problem formulation, we hypothesize that multiple viewpoints can benefit collision prediction since leveraging all viewpoints can better capture whole-body movement and environment context. 
This is confirmed by comparing the top section of \cref{table:ablation} (single-view) to the middle section (multi-view). We see a significant performance gain from using multi-view inputs, especially when combined with heatmap prediction, resulting in a $17.7\%$ and $12.4\%$ increase for \textit{F1@10} in training and unseen scenes, respectively.

\paragraph{Multi-modal support}
As \name can be easily extended to incorporate additional sensor modalities, \eg depth maps, we next test our model using depth as input instead of RGB.
As expected, this improves performance on both overall and per-joint predictions (bottom of \cref{table:ablation}).
This is expected because depth provides direct information on scene proximity, which is essential for our task.

\subsection{Collision Region Localization}
\label{subsec:qualitative}
\begin{figure*}[t]
    \centering
    \includegraphics[width=.96\linewidth]{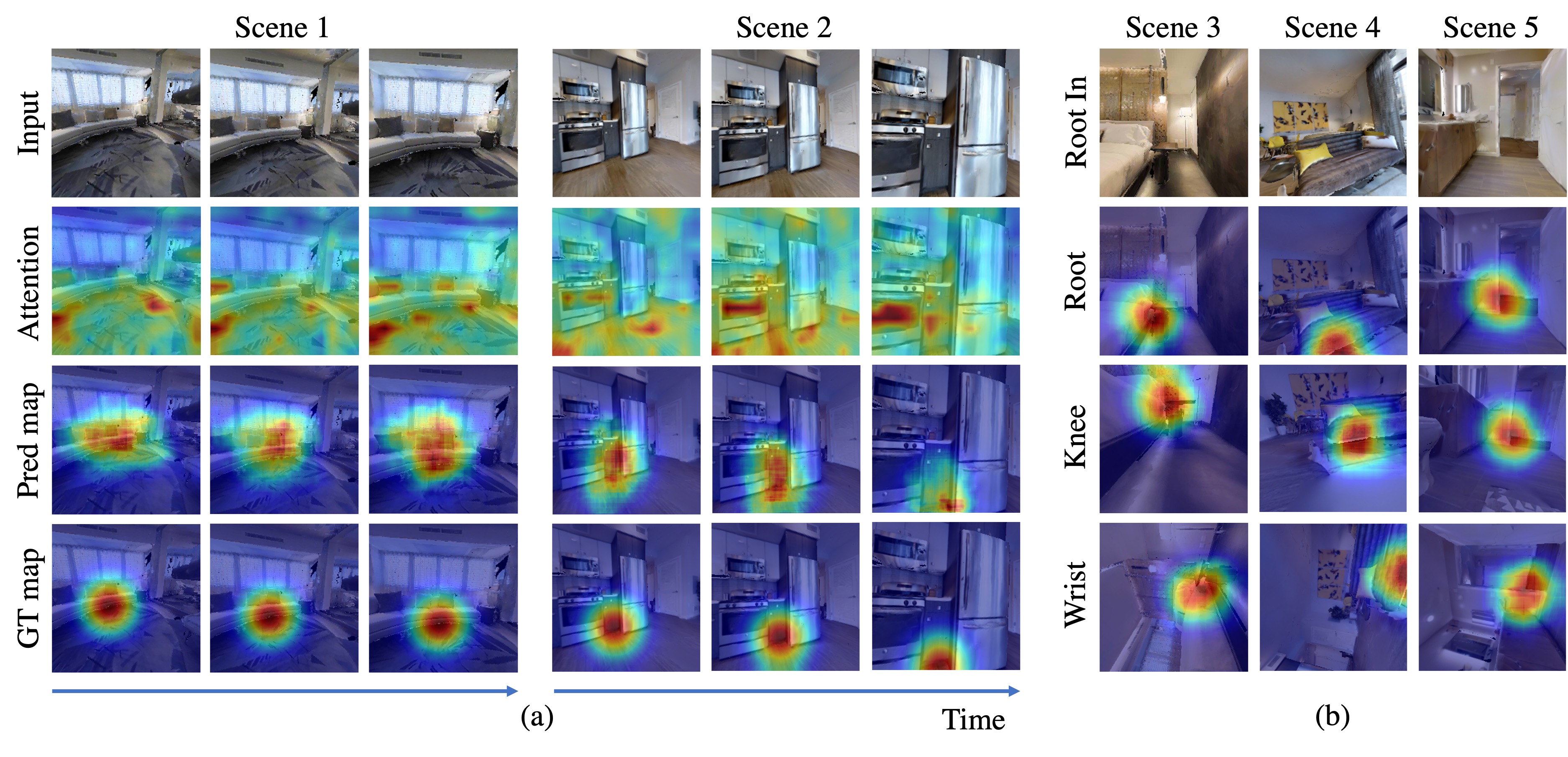}
    \vspace{-5mm}
    \caption{
    \textbf{Collision region localization results on unseen synthetic scenes.} \textbf{(a)} Two examples comparing \textit{\name Root-Only RGB} (\textit{Pred map}) to the ablated variant that does not predict heatmaps (\textit{Attention}). Explicit heatmap prediction localizes collision much more accurately. \textbf{(b)} Three examples from the full multi-view \name model. Predictions from multiple viewpoints are visualized at the same timestep. All examples are on the unseen scenes.
    }
    \vspace{-4mm}
    \label{fig:col_heatmap}
\end{figure*}

Next, we evaluate whether it is necessary to explicitly supervise collision region localization, or if meaningful spatial attention can emerge naturally through the prediction task itself. For this, we consider the single-viewpoint setting, and compare our \textit{\name Root-Only RGB} model to the ablated variant where no collision heatmap is predicted or supervised. We visualize the learned attention weights for this variant using Attention Rollout \cite{AttentionRollout2020}.
Multiple predicted collision region heatmaps are visualized in Fig.~\ref{fig:col_heatmap}~(a), where we show three sampled frames for each sequence. We observe that our model with direct heatmap supervision produces meaningful and concentrated heatmaps around reasonable risky regions, unlike the inherent attention learned without direct heatmap supervision. The heatmap offers important information about the potential collision event, \eg in \textit{Scene 2}, the left side of the oven is closer to the person, but only the right side is highlighted since the person moves in that direction.

In Fig.~\ref{fig:col_heatmap}~(b), we investigate if the predicted heatmaps are view-consistent. 
The same timestep is visualized from multiple viewpoints along with the collision region heatmaps from the full \name RGB model. The localization is consistent across viewpoints, suggesting that the model has learned a useful representation of the surrounding scene to enable spatial cross-view correspondences.

\begin{figure}[t]
    \centering
    \includegraphics[width=\linewidth]{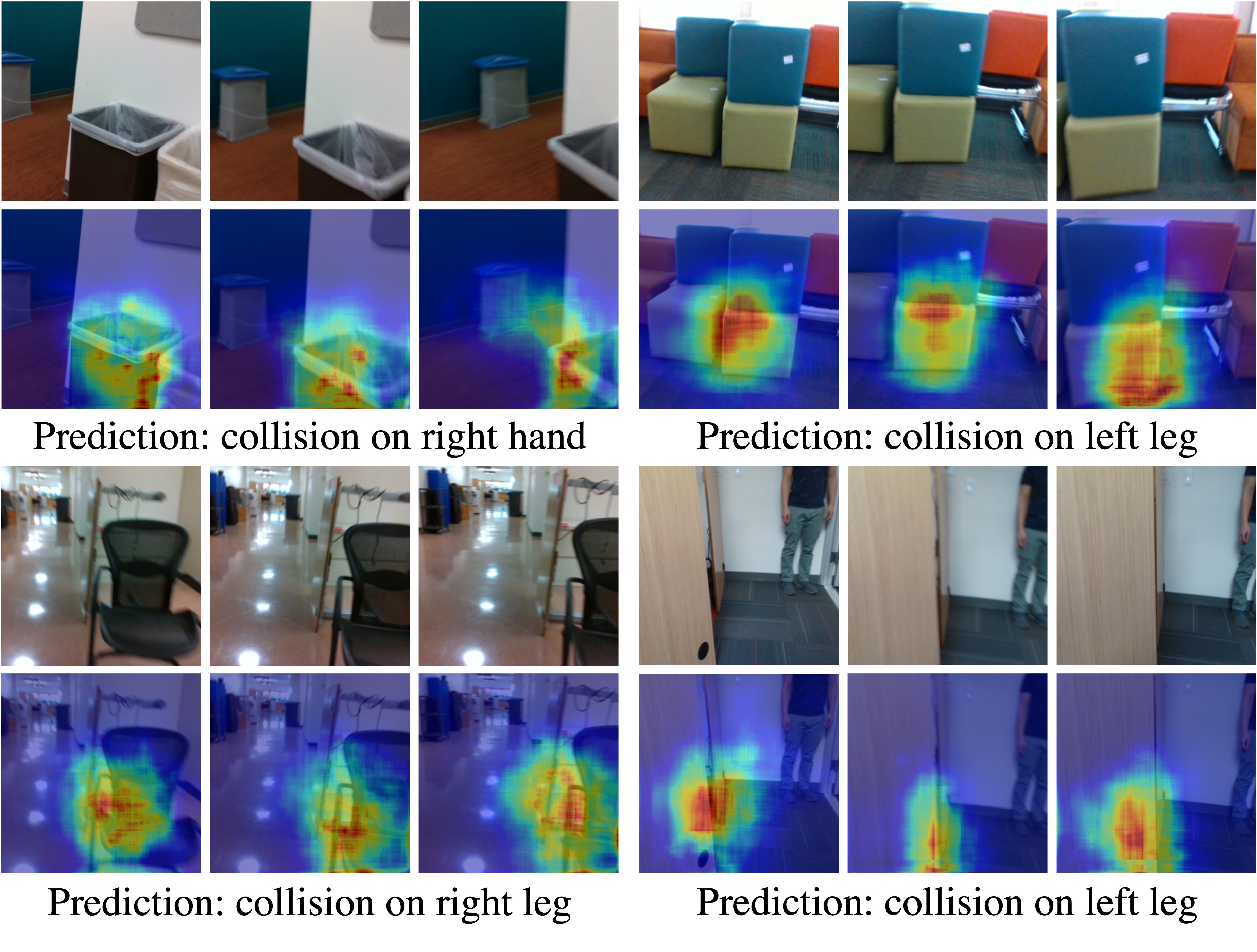}
    \vspace{-7mm}
    \caption{
    \textbf{Sim-to-real transfer.} Predictions from \textit{\name Root-Only RGB} for four real-world videos.
    }
    \vspace{-5mm}
    \label{fig:real_world}
\end{figure}

\subsection{Generalization to Real-World Videos}
\label{subsec:real_world_exp}
We show that our model trained on synthetic data can be directly used on real-world videos (see  Fig.~\ref{fig:problem_setup} \& \ref{fig:real_world}).
In Fig.~\ref{fig:real_world}, we show the results of applying the \textit{\name-Root-RGB} model to four real-world videos captured from a \textit{chest}-mounted camera as a person moves in cluttered offices. Our model reasonably localizes the risky regions that might cause collisions. For the first video (top left), as the person passes the trash bin on the right, our model predicts a potential collision with the right hand and the heatmap highlights the protruding object. We provide extensive visualizations in the supplementary video demonstrating \name's sim-to-real transfer ability, despite motion blur in the real-world videos and being trained on a different viewpoint.

\vspace{-1mm}
\subsection{Collision Avoidance Assistance}
\label{subsec:control_exp}
We additionally show that our model's predictions can be used to provide collision avoidance assistance to a virtual human in a simulated environment. Our setup is akin to a hip exoskeleton emulator with actuation to adjust foot placement and center-of-mass states~\cite{ExoMEppl2021}. We approximate the center of mass as the root joint and directly control its orientation.
Given the past 1 second of video observations, \textit{\name-Root-RGB} predicts if there will be a collision in the next 1 second. Based on the aggregated probabilities in the left and right halves of the predicted heatmap, the controller slightly rotates the hip by 5 degrees to the left or the right about the z-axis to avoid the higher-probability region.
Observations are then sampled at the new rotated pose and the control loop repeats autoregressively.
\begin{figure}
    \centering
    \includegraphics[width=\linewidth]{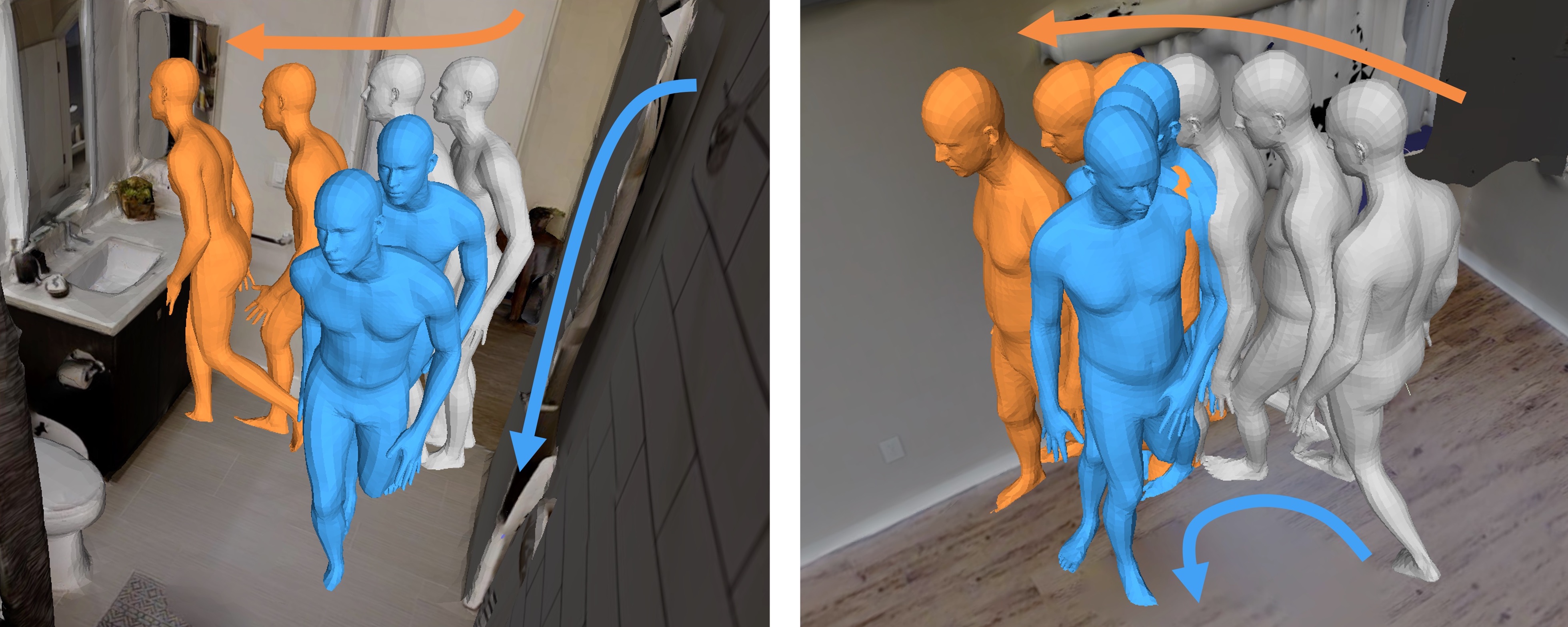}
\caption{
    \textbf{Collision avoidance visualization.} Grey meshes indicate the history trajectory. Orange is the original future without intervention, while blue uses collision avoidance assistance. The arrows show the direction of movement.
    }
    \vspace{-6mm}
    \label{fig:control_exp}
\end{figure}

The policy avoids 35\% of collision cases on training scenes and 29\% on unseen scenes. Two such cases are visualized in Fig.~\ref{fig:control_exp}. In the left example, the person originally walks into the vanity, but after the intervention, they instead gradually turn left to avoid the collision.

\section{Conclusion and Limitations}
In this work, we formulate the problem of collision prediction and localization from egocentric videos. To solve this problem, we propose a model named \name that leverages novel 4D space-time-viewpoint attention to aggregate motion and environment information from multiple viewpoints. We demonstrate that \name generalizes to unseen synthetic scenes as well as the real world, and provides useful information for collision avoidance.

Despite the promising results, our model has limitations.
Since generating scene-aware 3D human motion sequences is an open research problem~\cite{hassan2021stochastic, LongTermMotion2021, DiverseMotion2022,huang2023diffusion}, the motions in our dataset are scene-agnostic, which provide diverse collisions for training but do not always accurately reflect real-world scenarios. Moreover, our model struggles with multi-modal futures due to ambiguous motion patterns.
Finally, developing more sophisticated collision avoidance algorithms that leverage the detailed per-joint outputs of our model is another important direction.

\section{Acknowledgment}
This work was supported by a grant from the Stanford Human-Centered AI Institute (HAI), a Vannevar Bush Faculty Fellowship, an NVIDIA Graduate Fellowship, and the Swiss National Science Foundation under grant number P500PT 206946. We would like to thank colleagues from the Stanford Geometric Computation Group for valuable discussion and constructive feedback.

{\small
\bibliographystyle{ieee_fullname}
\bibliography{egbib}
}

\newpage
\clearpage
\appendix

In the supplement, we first provide additional visualizations (Sec.~\ref{supp:qualitative}) and quantitative analysis (Sec.~\ref{supp:quantitative}) to supplement the experiments in the main paper. We then follow with details of our implementation, including the model, training procedure, data generation, and experiments (Sec.~\ref{supp:implement_details}).

\section{Additional Visualizations}
\label{supp:qualitative}

\begin{figure*}
    \centering
    \includegraphics[width=\linewidth]{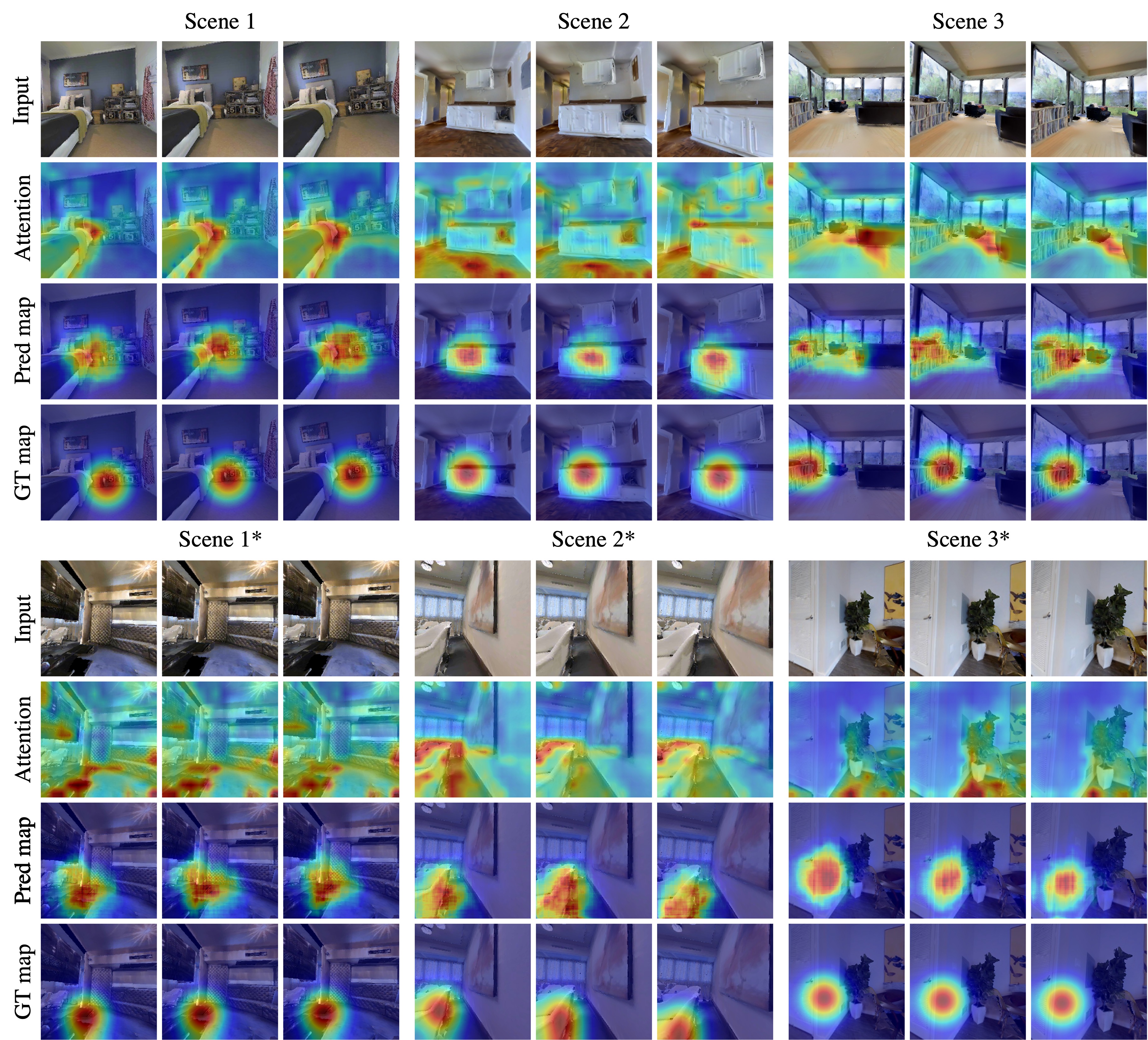}
    \caption{
    Additional collision region localization visualizations from \name Root-Only (\textit{Pred map}) with comparison to the ablated variant that does not predict heatmaps (\textit{Attention}). The top contains three examples sampled from the training scenes, while the bottom shows results from the unseen test scenes (*). Columns within each example progress in the temporal order.
    }
    \label{fig:addt_root_heatmap}
\end{figure*}

\parahead{Additional visualizations from \name Root-Only}%
We present additional collision region heatmap visualizations with \textit{\name Root-Only RGB}, which operates only on the pelvis viewpoint, in Fig.~\ref{fig:addt_root_heatmap}. \textit{\name Root-Only RGB} produces reasonable localization results on both the training and unseen scenes. 

\begin{figure*}
    \centering
    \includegraphics[width=\linewidth]{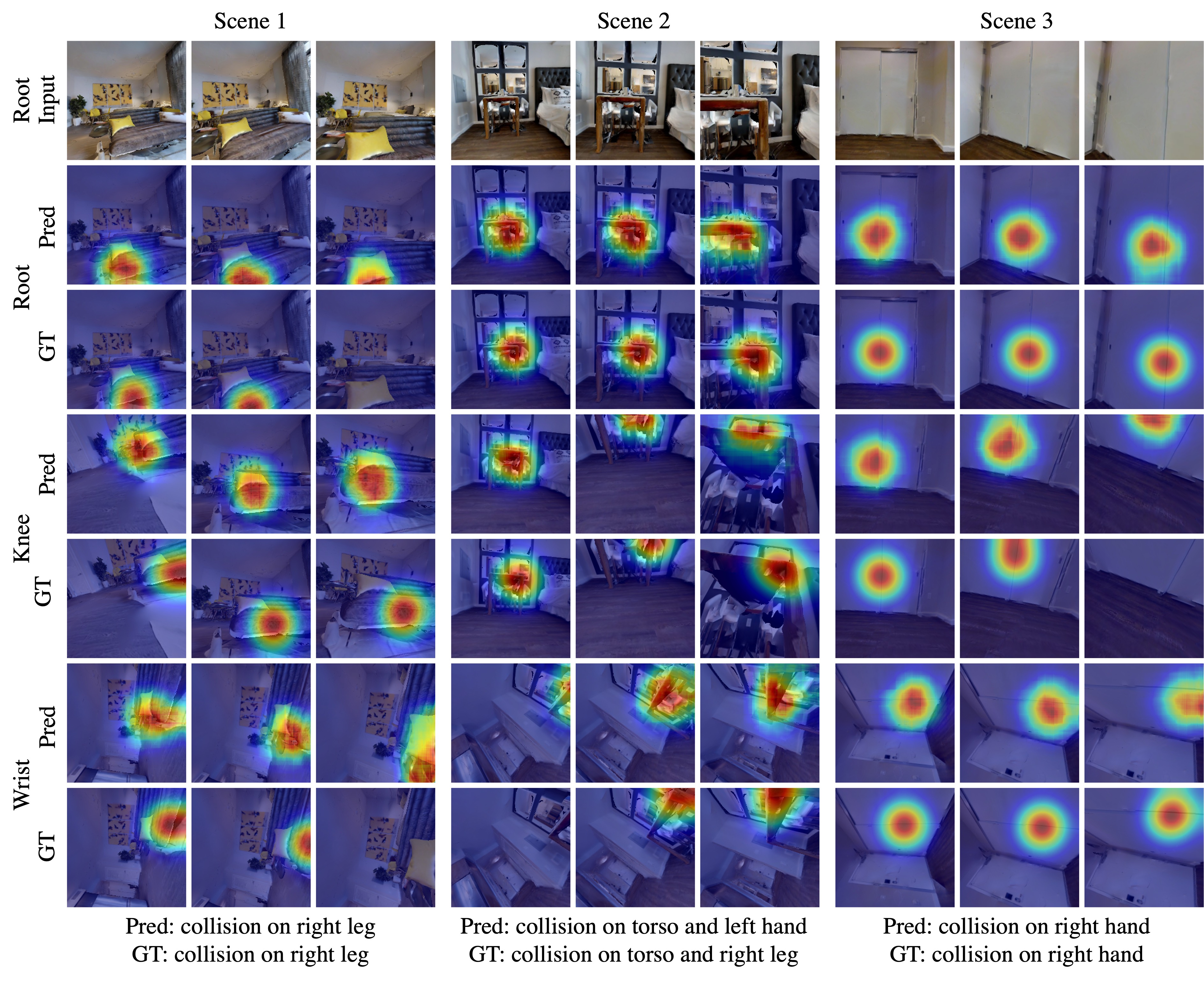}
    \caption{
    Additional collision region localization visualizations with collision predictions from the full multi-view \name model on unseen test scenes. Predicted and ground-truth heatmaps are shown from three of the six viewpoints. The model's prediction of the collision along with the ground-truth is shown at the bottom.
    }
    \label{fig:addt_all_vp_heatmap}
\end{figure*}

\parahead{Additional predictions from full \name}%
We show additional heatmap visualizations along with collision predictions from the full RGB \name model operating on all input views in Fig.~\ref{fig:addt_all_vp_heatmap}. \name generates collision region heatmaps that are consistent across viewpoints, without using explicit regularization on cross-viewpoint consistency. \name is also able to track the collision region across timesteps. This suggests that \name learns a good representation of the surrounding environment. \name also produces reasonable collision predictions. In the middle example of \cref{fig:addt_all_vp_heatmap}, the person gradually turns and walks to the right side of the nightstand, so \name predicts that the person is going to collide with their torso and left hand. Though in the ground truth, the right leg collides with the bed frame first, this is a hard example even for human observers; \name is still able to make a reasonable prediction.

\begin{figure}
    \centering
    \includegraphics[width=\linewidth]{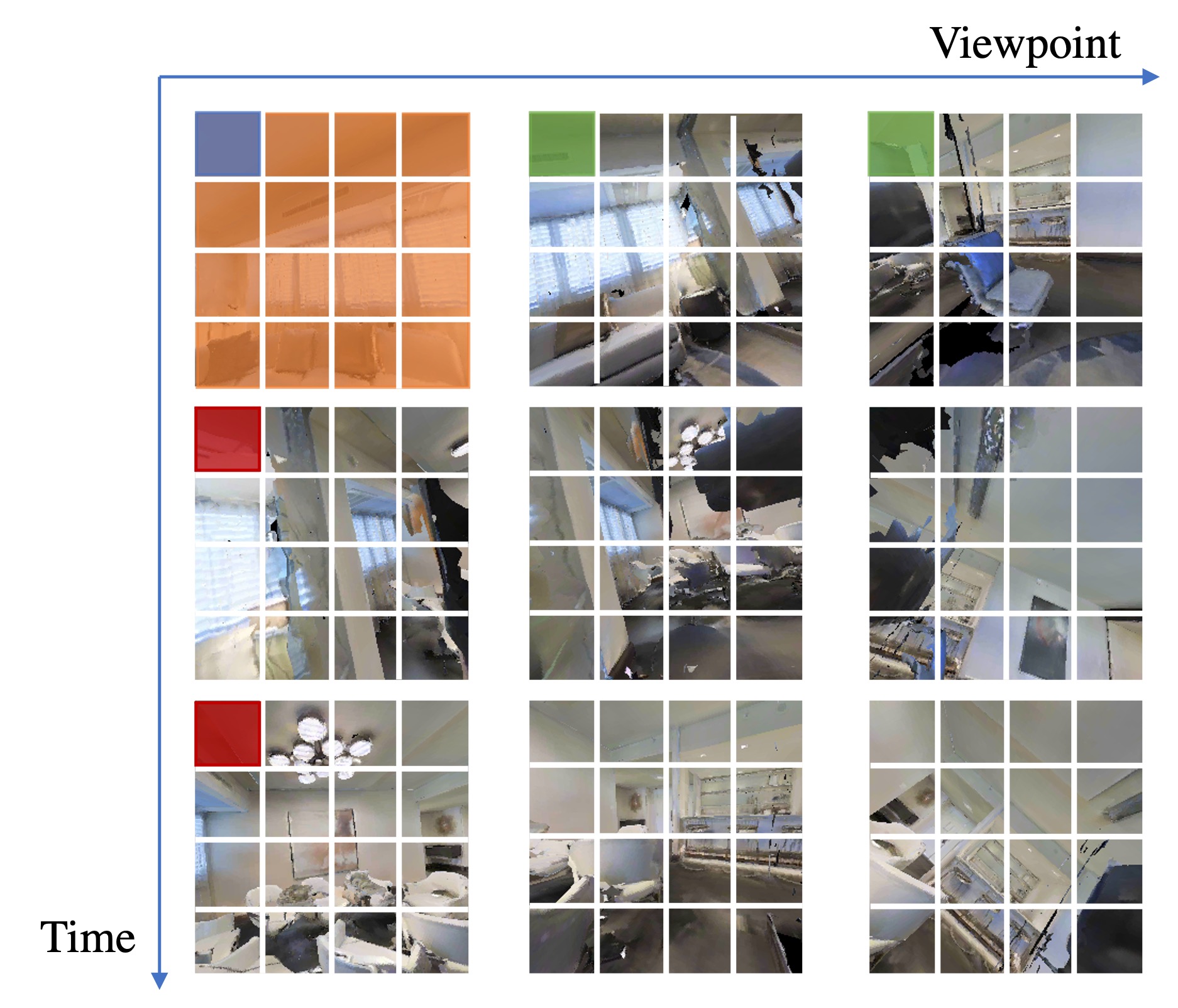}
    \caption{
    \textbf{Illustration of 4D Divided Space-Time-Viewpoint Attention (\textit{4D STV Div})}. The blue patch represents the query, while other patches with the same color denote its neighbors along each attention dimension.
    }
    \label{fig:attn_scheme}
    \vspace{-4mm}
\end{figure}

\begin{figure*}[ht!]
    \centering
    \includegraphics[width=\linewidth]{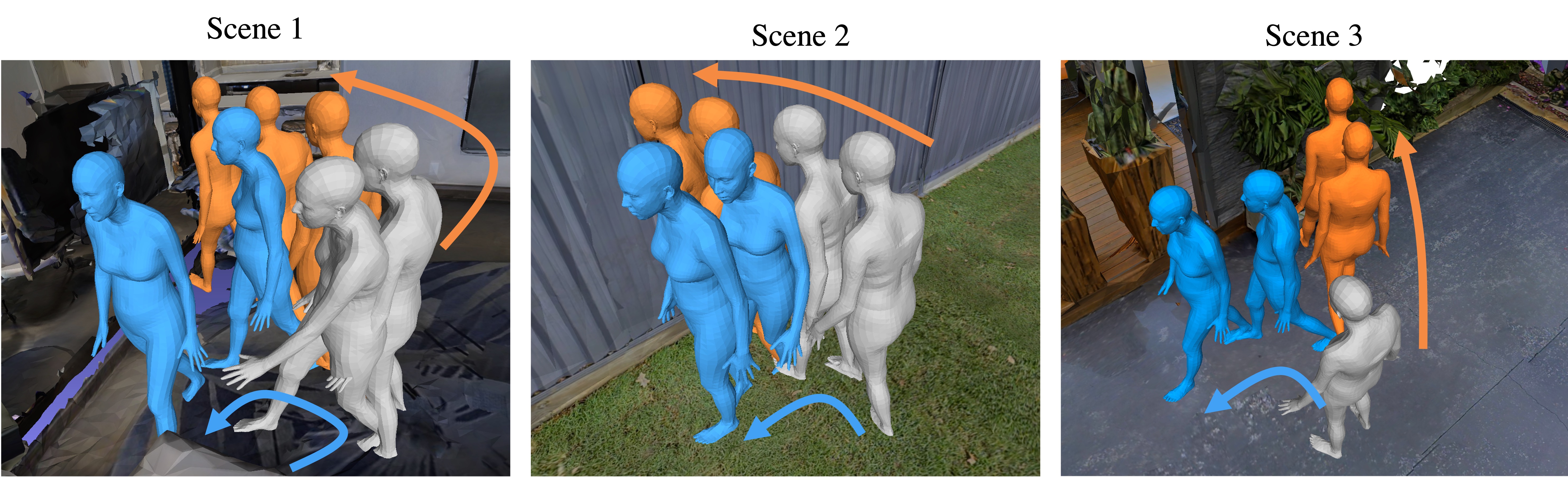}
    \caption{
    Collision avoidance assistance visualizations using control based on the output from \textit{\name Root-Only RGB}. Examples are shown for unseen test scenes. Grey meshes indicate the history trajectory. Orange is the original future without intervention, and blue is the future using collision avoidance assistance.}
    \label{fig:addt_control}
\end{figure*}

\parahead{Additional visualizations on collision avoidance}%
Following the setup introduced in \cref{subsec:control_exp} and Sec.~\ref{supp:control_exp}, we perform collision avoidance assistance on additional sequences and show the third-person visualizations in Fig.~\ref{fig:addt_control}. \name is able to provide useful signals that allow effective assistance, even in scenes not seen during training.

\begin{figure*}[ht!]
    \centering
    \includegraphics[width=\linewidth]{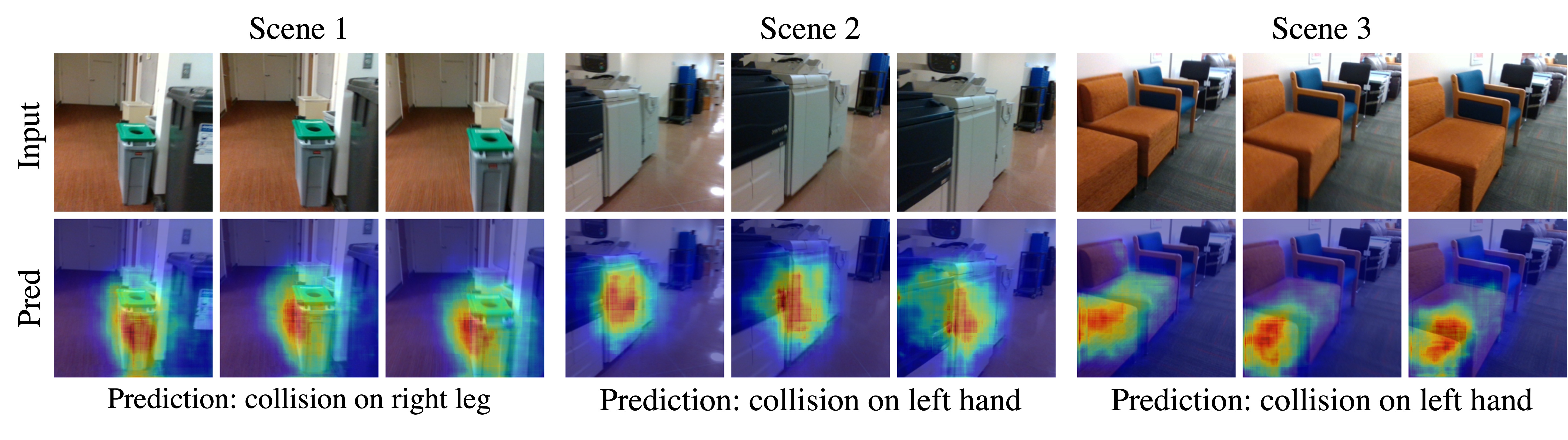}
    \vspace{-7mm}
    \caption{
    Additional prediction results on real-world videos from \textit{\name Root-Only RGB}.
    }
    \label{fig:addt_real_world}
\end{figure*}

\parahead{Additional predictions on real-world videos}%
Fig.~\ref{fig:addt_real_world} contains additional heatmaps and collision prediction results obtained by applying \textit{\name Root-Only RGB} to real-world videos. The model gives reasonable predictions despite the distribution shift from simulation to real-world and viewpoint change (these videos are captured from a chest-mounted camera, while training uses the root pelvis joint).

\section{Supplementary Quantitative Analysis}
\label{supp:quantitative}

\parahead{Complete results in all metrics}%
In \cref{table:quant_eval} and \cref{table:ablation} of the main manuscript, the quantitative results are only presented for F1 scores due to space limitations. Here, all results including precision, recall, and F1 scores at varying resolutions are shown in \cref{supp:base_prec,supp:base_recall,supp:ablation_f1,supp:ablation_prec,supp:ablation_rec}. %
Note that additional details on metrics are provided later in Sec.~\ref{supp:metrics}.

In \cref{supp:base_prec,supp:base_recall}, we see that our transformer backbone and proposed \textit{4D STV Full} attention scheme outperform all baselines in both precision and recall. 
In \cref{supp:ablation_f1,supp:ablation_prec,supp:ablation_rec} we see the same trends noted in the main paper for unseen scenes, namely multi-task and multi-view predictions improve performance, and that \name is flexible to the depth modality.

\parahead{Attention schemes}%
The original TimeSformer \cite{Timesformer2021} supports multiple attention schemes across space-time. In particular, the authors experimentally observe that the ``Divided Space-Time Attention'' scheme achieves the best balance between video classification performance and computational cost. This scheme only attends to the patches at the same spatial location across the time dimension, followed by a spatial self-attention to all patches in its own frame. We adapt this scheme to our 4D space-time-viewpoint space (``Divided Space-Time-Viewpoint attention'', or \textit{4D STV Div}) to compare with our proposed 4D attention scheme (\textit{4D STV Full}). \cref{fig:attn_scheme} illustrates the idea of the 4D divided attention scheme. Concretely, the query patch first attends to the patches at the same spatio-temporal location across all viewpoints, followed by patches at the same spatial location across all frames within its own viewpoint. Finally, it attends to all patches within its own frame. 

We compare with this variant experimentally in F1 scores in \cref{table:quant_eval}, and \cref{supp:base_prec} and \cref{supp:base_recall} in precision and recall scores, respectively. Across all metrics, our proposed space-time-viewpoint attention scheme (\textit{4D STV Full}) outperforms the Divided Space-Time-Viewpoint attention (\textit{4D STV Div}) baseline.

\begin{table*}[t!]
    \centering
    {%
    \setlength{\tabcolsep}{4pt}
    \begin{tabular}{lcl|ccc|ccc}
        \toprule
        & \multicolumn{1}{c}{} & \multicolumn{1}{c|}{} & \multicolumn{3}{c|}{\textbf{Unseen Motions}} & \multicolumn{3}{c}{\textbf{Unseen Motions \& Scenes}} \\
        \textbf{Method} & \textbf{Views} & \multicolumn{1}{c|}{\textbf{Attention}} & Prec@4 & Prec@6 & Prec@10 & Prec@4 & Prec@6 & Prec@10 \\
        \midrule
        Time2Col~\cite{KitaniCollision2019} & Root & \multicolumn{1}{c|}{-} & 56.1 & 43.5 & \textbf{38.3} & 47.7 & 34.1 & 29.9 \\
        \name & Root & 3D ST \cite{Timesformer2021} & \textbf{57.1} & \textbf{44.2} & 37.6 & \textbf{55.9} & \textbf{40.2} & \textbf{32.9} \\
        \midrule
         &  & 3D ST \cite{Timesformer2021} & 66.3 & 51.0 & 45.9 & 52.8 & 37.7 & 33.2 \\
        \name & All & 4D STV Div & 65.5 & 55.8 & 50.9 & 52.2 & 42.4 & 35.6 \\
        &  & 4D STV Full (ours) & \textbf{68.6} & \textbf{58.7} & \textbf{51.5} & \textbf{59.6} & \textbf{48.6} & \textbf{41.1} \\
        \bottomrule
    \end{tabular}
    }
    \caption{Comparison with baselines in precision scores (\%).}

    \label{supp:base_prec}
\end{table*}

\begin{table*}[t!]
    \centering
    {%
    \setlength{\tabcolsep}{4pt}
    \begin{tabular}{lcl|ccc|ccc}
        \toprule
        & \multicolumn{1}{c}{} & \multicolumn{1}{c|}{} & \multicolumn{3}{c|}{\textbf{Unseen Motions}} & \multicolumn{3}{c}{\textbf{Unseen Motions \& Scenes}} \\
        \textbf{Method} & \textbf{Views} & \multicolumn{1}{c|}{\textbf{Attention}} & Rec@4 & Rec@6 & Rec@10 & Rec@4 & Rec@6 & Rec@10 \\
        \midrule
        Time2Col~\cite{KitaniCollision2019} & Root & \multicolumn{1}{c|}{-} & \textbf{61.0} & 48.2 & 41.7 & 50.2 & 36.4 & 32.1 \\
        \name & Root & 3D ST \cite{Timesformer2021} & 60.9 & \textbf{48.9} & \textbf{42.2} & \textbf{60.6} & \textbf{45.5} & \textbf{36.8} \\
        \midrule
         &  & 3D ST \cite{Timesformer2021} & 77.8 & 67.8 & 60.8 & 69.2 & 55.3 & 48.5 \\
        \name & All & 4D STV Div & 73.2 & 64.6 & 59.2 & 60.6 & 49.9 & 43.8 \\
        &  & 4D STV Full (ours) & \textbf{81.8} & \textbf{72.5} & \textbf{67.8} & \textbf{75.7} & \textbf{64.8} & \textbf{58.6} \\
        \bottomrule
    \end{tabular}
    }
    \caption{Comparison with baselines in recall scores (\%).}

    \label{supp:base_recall}
\end{table*}

\begin{table*}[t!]
    \centering
    {%
    \setlength{\tabcolsep}{2pt}
    \begin{tabular}{ccc|ccc|ccc}
        \toprule
        \multicolumn{2}{c}{\textbf{\small Input}} & \textbf{\small Output} & \multicolumn{3}{c}{\textbf{\small Unseen Motions}} & \multicolumn{3}{c}{\textbf{\small Unseen Motions \& Scenes}} \\
        Views & Modality & Map & F1@4 & F1@6 & F1@10 & F1@4 & F1@6 & F1@10\\
        \midrule
        Root & RGB & \xmark & \textbf{59.9} & \textbf{44.9} & \textbf{38.7} & 50.7 & 39.1 & 32.4 \\
        Root & RGB & \cmark & 57.1 & 44.7 & 37.9 & \textbf{56.0} & \textbf{40.7} & \textbf{33.2} \\
        \midrule
       All & RGB & \xmark & 59.1 & 48.7 & 43.0 & 50.9 & 39.2 & 33.0 \\
        All & RGB & \cmark & \textbf{71.9} & \textbf{62.2} & \textbf{55.6} & \textbf{63.7} & \textbf{52.9} & \textbf{45.6} \\
        \midrule
       All & Depth & \cmark & 76.3 & 71.1 & 64.7 & 72.6 & 66.3 & 60.1 \\
        \bottomrule
    \end{tabular}
    }
    \caption{Ablation study in F1 scores (\%).}

    \label{supp:ablation_f1}
\end{table*}

\begin{table*}[t!]
    \centering
    {%
    \setlength{\tabcolsep}{2pt}
    \begin{tabular}{ccc|ccc|ccc}
        \toprule
        \multicolumn{2}{c}{\textbf{\small Input}} & \textbf{\small Output} & \multicolumn{3}{c}{\textbf{\small Unseen Motions}} & \multicolumn{3}{c}{\textbf{\small Unseen Motions \& Scenes}} \\
        Views & Modality & Map & Prec@4 & Prec@6 & Prec@10 & Prec@4 & Prec@6 & Prec@10\\
        \midrule
        Root & RGB & \xmark & \textbf{58.5} & 42.7 & 36.8 & 49.1 & 36.9 & 31.1 \\
        Root & RGB & \cmark & 57.1 & \textbf{44.2} & \textbf{37.6} & \textbf{55.9} & \textbf{40.2} & \textbf{32.9} \\
        \midrule
       All & RGB & \xmark & 54.3 & 43.7 & 37.9 & 46.2 & 34.2 & 28.8 \\
        All & RGB & \cmark & \textbf{68.6} & \textbf{58.7} & \textbf{51.5} & \textbf{59.6} & \textbf{48.6} & \textbf{41.1} \\
        \midrule
       All & Depth & \cmark & 72.6 & 66.9 & 60.3 & 69.2 & 62.3 & 54.6 \\
        \bottomrule
    \end{tabular}
    }
    \caption{Ablation study in precision scores (\%).}
    \label{supp:ablation_prec}
\end{table*}

\begin{table*}[t!]
    \centering
    {%
    \setlength{\tabcolsep}{2pt}
    \begin{tabular}{ccc|ccc|ccc}
        \toprule
        \multicolumn{2}{c}{\textbf{\small Input}} & \textbf{\small Output} & \multicolumn{3}{c}{\textbf{\small Unseen Motions}} & \multicolumn{3}{c}{\textbf{\small Unseen Motions \& Scenes}} \\
        Views & Modality & Map & Rec@4 & Rec@6 & Rec@10 & Rec@4 & Rec@6 & Rec@10\\
        \midrule
        Root & RGB & \xmark & \textbf{66.7} & \textbf{52.0} & \textbf{45.2} & 57.4 & 46.1 & 38.3 \\
        Root & RGB & \cmark & 60.9 & 48.9 & 42.2 & \textbf{60.6} & \textbf{45.5} & \textbf{36.8} \\
        \midrule
       All & RGB & \xmark & 71.9 & 62.3 & 56.8 & 63.7 & 51.8 & 45.0 \\
        All & RGB & \cmark & \textbf{81.8} & \textbf{72.5} & \textbf{67.8} & \textbf{75.7} & \textbf{64.8} & \textbf{58.6} \\
        \midrule
       All & Depth & \cmark & 86.6 & 82.3 & 78.0 & 84.1 & 79.2 & 76.2 \\
        \bottomrule
    \end{tabular}
    }
    \caption{Ablation study in recall scores (\%).}
    \label{supp:ablation_rec}
\end{table*}

\section{Implementation Details}
\label{supp:implement_details}
\subsection{\name Architecture}
As opposed to the original TimeSformer \cite{Timesformer2021}, we only use 8 layers of attention blocks, with 8 attention heads each and the model gets 5 video frames (1 second at 5 Hz) for the input. We initialize our network with the publicly available weights of TimeSformer pretrained on Kinetics-600~\cite{Kinetics2018}. For experiments that use depth as input, we simply sum the pre-trained weights across the three input channels for the first convolution layer to initialize the network.

\subsection{Supervision on Each Sub-task}
We compute the overall collision prediction loss $L_{\text{col}}$ and the collision region localization loss $L_{\text{map}}$ on all training examples, while only computing the per-joint collision prediction loss $L_{\text{joint}}$ on training examples with $y^{\text{col}} = 1$ (\ie examples where a collision happens in the ground truth). This is because the majority ($\sim$70\%) of samples in our dataset do not contain a collision incident, and for those that contain a collision, only a few joints are colliding, which complies with the real-world data distribution. If we supervise the joint-level collision prediction on all examples, the model will be overwhelmed with gradients computed from the negative samples, and hence always predicts no joints colliding. Hence, we factorize the collision prediction problem into the sub-problems of overall and joint-level collision prediction, which makes the optimization less susceptible to excessive negative samples. At test time, the overall collision prediction can be used as a gate to determine if the per-joint collision and heatmap predictions should be used or not, \ie they can be ignored if the overall collision prediction is negative since no collision avoidance action is needed.

\subsection{Data Generation}
In this section, we provide full details of our data generation pipeline as well as the benchmark dataset that we use throughout our experiments.

\parahead{Photo-realistic synthetic scenes} To enable cross-scene generalization by our learned model, it is crucial to have a large variety of scenes in training. We pick 100 scenes from the Matterport3D \cite{Matterport3d2017} and Gibson \cite{Gibson2018} datasets that have a traversable area between 0 and 150$\text{m}^2$ and each contains several rooms. Note that our data infrastructure is agnostic to the scenes used, and it is trivial to add any other 3D scene datasets available.

\parahead{Human motion generation} We leverage HuMoR \cite{Humor2021} to generate realistic and diverse human motions within each scene. To encourage generating walking motions that are most common, we train HuMoR on a subset of the large-scale motion capture dataset AMASS \cite{AMASS2019} containing mostly walking motions by using the motion labels from BABEL \cite{BABEL2021}. HuMoR generates motion for a random human body for $T_{\text{eval}} = 300$ time steps at 30 Hz and collision checking (see below) is performed at every step. If a collision happens, generation is stopped and the sequence is saved if its length is longer than a threshold $T_{\text{min}} = 30$, otherwise it is discarded. If no collision happens, the entire sequence is saved. This results in a set of motion sequences $M=\{m_1, m_2, \ldots, m_{N_\text{orig}}\}$, where each $m_n$ has a temporal length of $t_n$, $T_{\text{min}} \leq t_n \leq T_{\text{eval}}$. Since each synthetic scene comes with a different area, the number of motion sequences generated in each varies from $N_{\text{min}} = 50$ to $N_{\text{max}} = 250$. To exclude backward motions, we additionally check if the root orientation and root velocity have a dot product larger than 0.3 for the last 30 time steps.

\parahead{Collision checking and label generation} The goal of this step is to identify which of $J = 10$ human body joints (head, torso, elbows, hands, legs, and feet) collide with the environment when a collision happens. We first segment the human body into parts with the SMPL body segmentation map \cite{SMPL2015}. Convex hulls are created for each part and used to perform collision checking with the scene mesh via PyBullet \cite{PyBullet2021}. At the time step of collision, we compute the Euclidean distance of each colliding body vertex to all 10 body joint locations, and take the closest joint as the colliding joint label $y^{\text{joint}}$. The overall binary collision label $y^{\text{col}}$ is obtained by checking if any of the joints are in a collision. In order to avoid false positives caused by faulty floor penetration from HuMoR, we remove the floor when checking for collisions.

\parahead{Post-processing of motion sequences} Initial motion sequences $M$ are next split into equal length sub-sequences of size $T_{\text{orig}} = 30$ to be used as model input. We use a sliding window strategy with the stride size being $S = 30$ to get a set of sub-sequences $M' = \{m'_1, m'_2, \ldots, m'_{N_\text{sub}}\}$. Each sub-sequence is labeled as leading to a collision if a collision occurs within the following $H = 30$ frames (the prediction horizon). 

\parahead{Observation generation} From the processed sub-sequences, we retrieve the translation and orientation vectors for $V = 6$ body joints: head, pelvis, wrists, and knees. These poses are used to place cameras at each joint and render egocentric observations within the AI-Habitat simulator \cite{Habitat2021,Habitat2019}. These image sequences are then sub-sampled uniformly from $T_{\text{orig}}$ to $T = 5$ frames, which are the model inputs.  

\parahead{Collision region heatmap generation} When a collision happens, we record the vertices on the scene mesh that collide with the human body. We then perform perspective projection to project these 3D points into the 2D egocentric image plane. If any of the points are visible in the image frame, we set the values of the collision region heatmap at those pixel locations $p^v_{tij}$ to 1 and leave other locations at 0. We then apply a Gaussian kernel to the resulting image for smoothening and normalize to sum to 1 to get a spatial distribution. If none of the points are visible, all values in the heatmap are set to 0. 

\subsection{Experiment Details}
\label{supp:exp_details}

\subsubsection{Evaluation Metrics}
\label{supp:metrics}
In this paper, we report the precision, recall, and F1 scores at different ``resolutions'' $N$. This means we split the joints into $N$ groups, and a group is considered colliding if \textit{any} joint in that group is colliding. Specifically, we have $N \in \{10, 6, 4\}$. When $N = 10$, this is the original evaluation setup, where every joint predicted by \name is evaluated on its own. When $N = 6$, we split the 10 joints into 6 groups based on their semantic proximity on \textit{each} side of the body, \ie [``head"], [``pelvis"], [``right hand", ``right elbow"], [``left hand", ``left elbow"], [``right foot", ``right leg"], and [``left foot", ``left leg"]. When $N = 4$, the joints are further clustered on \textit{both} sides of the body, \ie [``head"], [``pelvis"], [``right hand", ``right elbow", ``left hand", ``left elbow"], and [``right foot", ``right leg", ``left foot", ``left leg"]. 

The reason why we have multiple resolutions is that we may care about different granularities of joint-level predictions for different application scenarios. 
For example, if a person will collide with the right side of the body, it may be sufficient to know that the collision will be somewhere on the right arm without knowing whether it is the hand or elbow. 
Moreover, if a person will walk straight into a wall, it is difficult (and not very useful) to determine whether the left or right foot will collide first in stride, so it is sensible to evaluate prediction accuracy while ignoring these symmetries (as is the case for $N=4$).

\subsubsection{Additional Details on Baselines}
As mentioned in Sec.~\ref{subsubsec:attn}, there is no prior work directly comparable to ours since we study a new problem formulation. TimeSformer~\cite{Timesformer2021} and Time2Col~\cite{KitaniCollision2019} are the two closest works to ours for comparison. For a fair comparison, we extended both to perform our collision prediction and localization tasks. Specifically, TimeSformer takes a video from a single viewpoint, and performs spatio-temporal self-attention to extract the features, which are finally used for video classification. To adapt TimeSformer for our setting, we discard the first classification token $\mathbf{e}^{(L)}_{(0, 0, 0)}$ \cite{Timesformer2021}, and only use the remaining feature map $\mathbf{e}^{(L)}_{(1:, 1:, 1:)}$. We then concatenate our collision prediction and localization branches after this feature map.

Time2Col, on the other hand, adopts convolution-based approaches. The authors experimented with several video architectures for their task of predicting time-to-collision, including Single-Image VGG-16, I3D, and Multi-stream VGG. The best-performing variant was reported to be the Multi-stream VGG, which leverages per-frame 2D convolutions and concatenates the flattened frame-wise features for the final prediction. Similar to TimeSformer, we take the concatenated feature maps, and pass them to a two-layer MLP for collision prediction. For the collsion heatmaps, we use the un-flattened features instead, and append our localization branch after them. 

The training configurations for baselines are tuned separately to reach their best performance, and the best-performing checkpoints are used to report the results.

\subsubsection{Collision Avoidance Assistance Using \name}
\label{supp:control_exp}
We show the details of how we perform the collision avoidance assistance using \textit{\name Root-Only RGB} in 
Algorithm~\ref{alg:col_avoid} (notations follow Sec.~\ref{subsec:col_region_method} and Sec.~\ref{subsec:col_prediction_method}). A subtlety here is that throughout training and all other experiments, we uniformly sample the 30 input frames to 5 frames, \ie the 1st, 6th, 12th, 18th, and 24th frames. However, we perform autoregressive control assistance in this control experiment, which requires the observations to be updated at each time step in the future. Hence, we modify the sampling strategy here to use the 7th, 12th, 18th, 24th, and 30th frames, so that each new incoming frame gets appended to the tail of the frame stack.

\begin{algorithm}
    \caption{Control policy for collision avoidance assistance}
    \label{alg:col_avoid}
    \SetKwInOut{Input}{Input}
    \SetKwInOut{Output}{Output}
    \Input{Observation video $x$, rotation angle $\delta$}
    \Output{Control torques}
    $t \gets 0 $ \\
    $\hat{y}^{\text{col}}, \hat{y}^{\text{map}} \gets f(x)$ \\
    \While{$\hat{y}^{\text{col}} = 1$}{
        Rotate($\hat{y}^{\text{map}}$, $\delta$) \\
        $x \gets \text{GetNewObservations}(t) $ \\
        $\hat{y}^{\text{col}}, \hat{y}^{\text{map}} \gets f(x)$ \\
        $t \gets t + 1 $ \\
    }
    Where Rotate($\hat{y}^{\text{map}}$, $\delta$) is a function that rotates the human body by $\delta$ degrees in the opposite direction (left / right) of the side that has a larger predicted collision region in the heatmap. And $\text{GetNewObservations}(t)$ returns the new stack of sampled frames ending at time step $t$ following the procedure stated above.
\end{algorithm}

\subsubsection{Runtime Analysis}
In order to be deployed in real-world applications, it is important that \name can be run in real-time.
To this end, we evaluate \name inference time on an Nvidia A5000 24 GB GPU. Our full \name model takes on average 0.12 seconds to process a single example (\ie 6 view, 1 second RGB videos), while the \textit{\name Root-Only} variant spends only 0.02 seconds (\ie 1 view, 1 second RGB videos). This efficiency is enough for VR / AR or exoskeleton systems to react in real-time.

\end{document}